\documentclass[letterpaper]{article} 
\usepackage{aaai2026}  
\usepackage{times}  
\usepackage{helvet}  
\usepackage{courier}  
\usepackage[hyphens]{url}  
\usepackage{graphicx} 
\usepackage{natbib}  
\usepackage{caption} 
\usepackage{booktabs} 
\usepackage{algorithm}
\usepackage{algorithmic}

\usepackage[most]{tcolorbox}

\usepackage{newfloat}
\usepackage{listings}
\DeclareCaptionStyle{ruled}{labelfont=normalfont,labelsep=colon,strut=off} 
\floatstyle{ruled}
\newfloat{listing}{tb}{lst}{}
\floatname{listing}{Listing}
\title{Do Reasoning Models Ask Better Questions?\\A Formal Information-Theoretic Analysis on Multi-Turn LLM Games}
\author {
    Daniel M. Pedrozo\textsuperscript{\rm 1 2},
    Telma W. de L. Soares\textsuperscript{\rm 1 2},
    Bryan L. M. de Oliveira\textsuperscript{\rm 1 2}
}
\affiliations {
    \textsuperscript{\rm 1} Advanced Knowledge Center for Immersive Technologies – AKCIT, Brazil \\
    \textsuperscript{\rm 2} Institute of Informatics, Federal University of
Goiás, Goiânia, Brazil \\
    danielpedrozo@discente.ufg.br, telma\_woerle@ufg.br, bryanlincoln@discente.ufg.br
}

\begin{document}

\maketitle

\begin{abstract}
Large Language Models (LLMs) excel at many tasks but still struggle with a critical ability for LLM-based agents: asking good questions for resolving ambiguity in user requests. While prior work has explored information-seeking behavior through word games, existing benchmarks lack comprehensive evaluation frameworks that provide both final and intermediate signals based on Information Gain (IG). Moreover, they rarely provide systematic comparison between models that use Chain-of-Thought reasoning and those that do not. We propose a multi-turn dialogue framework that quantitatively measures how effectively LLMs gather information through yes/no questions in a hierarchical knowledge graph environment. Our framework employs a triad of interacting LLM agents that ask questions, answer them, and update the hypothesis space. We adopt IG as the main metric, grounded in Shannon entropy, to assess query effectiveness at each turn and cumulatively. We instantiate our framework in a geographical \emph{Guess My City} game setting organized in a five-level taxonomy and evaluate multiple LLM variants under fully and partially observable conditions, with and without Chain-of-Thought reasoning. Our experiments demonstrate that, among the evaluated models, the ones with explicit reasoning capabilities achieve higher IG per turn and reach solutions in fewer steps, particularly in partially observable settings. Analysis of reasoning traces reveals that smaller models compensate for limited capacity through more aggressive exploration of candidate questions, while larger models exhibit higher assertiveness in selecting optimal queries, generating candidates with greater potential IG.
\end{abstract}

\begin{links}
    \link{Code}{https://github.com/machadoDaniels/info-gainme}
\end{links}

\section{Introduction}
\begin{figure}[t]
\centering
\includegraphics[width=\columnwidth]{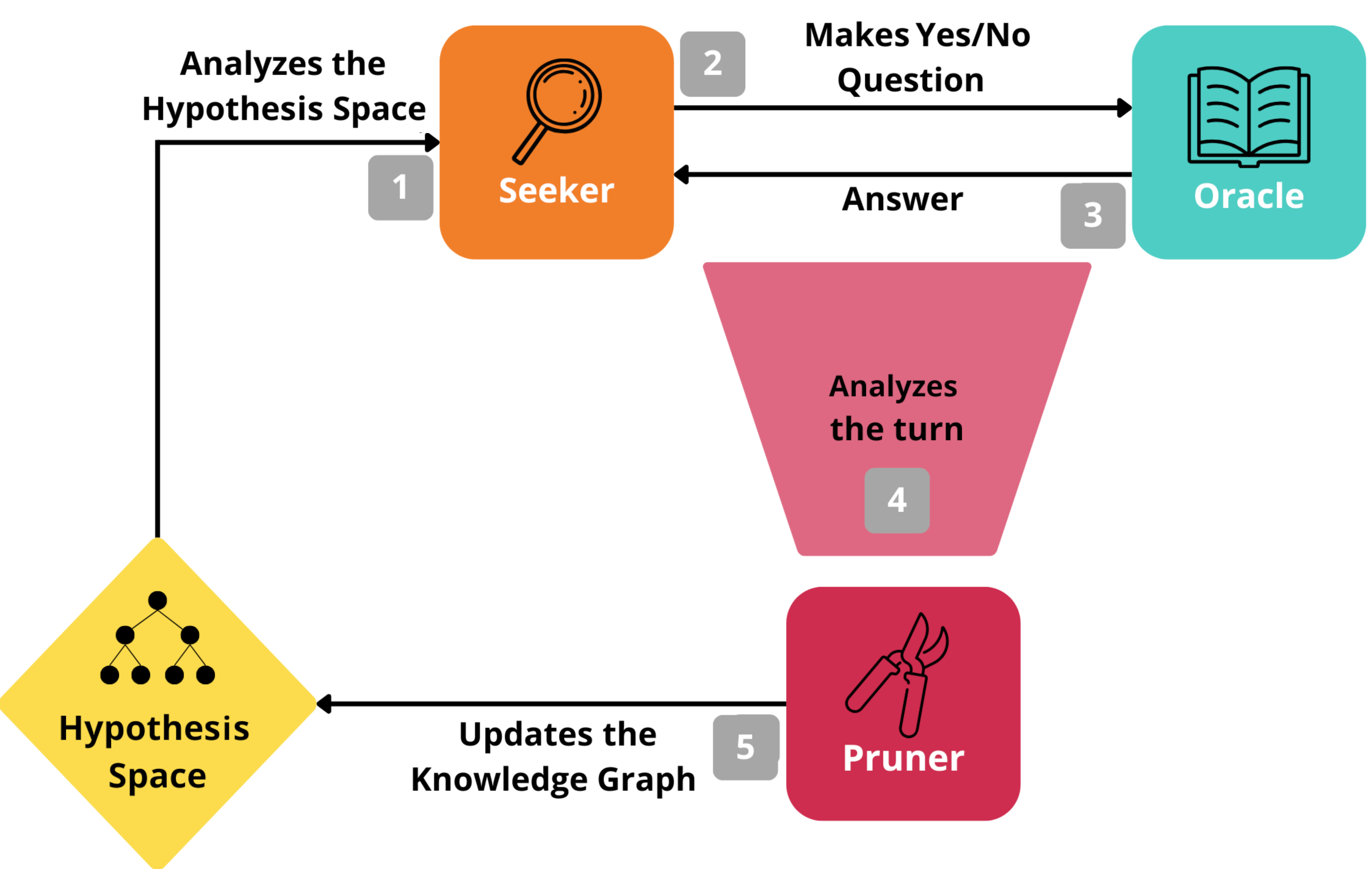}
\caption{\textbf{Framework architecture.} The numbered arrows indicate the sequential flow between the three agents and the hypothesis space.}
\label{fig:flow}
\end{figure}


Large language models have proven to be highly effective as chatbots, assistants, and in a variety of domains~\cite{grattafiori2024llama3herdmodels,openai2024gpt4technicalreport,yang2025qwen3technicalreport}. However, they continue to face significant challenges in recognizing and managing ambiguity in user requests, frequently responding only partially, either by selecting a single plausible interpretation or by attempting to address all conceivable meanings that may emerge from the ambiguity~\cite{deoliveira2025infoquestevaluatingmultiturndialogue, kuhn2023clamselectiveclarificationambiguous, abdulhai2023lmrlgymbenchmarksmultiturn}. This limitation becomes more prominent in real-world scenarios requiring both comprehensive understanding and active disambiguation, particularly where the environment is partially observable and uncertain~\cite{wu2025collabllmpassiverespondersactive, deoliveira2025infoquestevaluatingmultiturndialogue}.

To effectively navigate such uncertainties, we posit that a proficient agent must demonstrate: 
\begin{enumerate}
    \item The ability to recognize when a user query is ambiguous or when insufficient information is available to resolve it.
    \item Competence in estimating the space of hypotheses, evaluating the potential Information Gain (IG) associated with each, and selecting the question whose answer is expected to provide the highest reduction in uncertainty, which requires robust reasoning and planning abilities.
    \item The capacity to formulate questions that maximize IG, thereby reducing the set of plausible hypotheses.
\end{enumerate}

Prior work has explored information-seeking behavior through word games and structured hypothesis spaces. Notable approaches include LMRL-Gym~\cite{abdulhai2023lmrlgymbenchmarksmultiturn}, which uses word games as benchmarks for multi-turn tasks, training intrinsically curious LLMs~\cite{tajwar2025traininggenerallycuriousagent}, and leveraging Expected Information Gain in \emph{20-Questions} games~\cite{mazzaccara2024learningaskinformativequestions, bertolazzi-etal-2023-chatgpts}. However, these approaches either lack fine-grained IG metrics throughout the interaction or restrict their analyses to single models without examining the impact of Chain-of-Thought (CoT) reasoning on information-seeking performance.

To address these gaps, we propose a novel framework (see Figure~\ref{fig:flow}) for evaluating LLMs in multi-turn information-gathering tasks, where datasets are structured as explicit hypothesis spaces with fully enumerated candidate outcomes. Our framework provides granular, turn-level and final task feedback via a system that computes rewards through IG for each assistant-generated question. Generalizable to datasets from different domains for which the set of possible targets can be enumerated, this design enables a quantitative assessment of LLMs' information-seeking capabilities.

This work makes three main contributions. 
(i) We introduce a multi-turn evaluation framework for quantitatively assessing the information-seeking abilities of LLMs in structured hypothesis spaces under fully observable and partially observable conditions. 
(ii) We provide a systematic comparison of LLMs with and without Chain-of-Thought reasoning. 
(iii) We conduct a fine-grained analysis of reasoning traces to characterize how models of different sizes explore the hypothesis space and select informative questions. Together, these contributions enable a principled, information-theoretic perspective on question-asking behavior in LLM-based agents.

\section{Related Work}

Several prior works have attempted to tackle information-seeking challenges using different approaches, many of which rely on text games. LMRL-Gym~\cite{abdulhai2023lmrlgymbenchmarksmultiturn} leverages word games as benchmarks to evaluate agents on multi-turn tasks, and compares different LLMs by using the game's successes as a reward signal during training. \citet{tajwar2025traininggenerallycuriousagent} use Direct Preference Optimization and Supervised Fine-Tuning to train intrinsically curious LLMs using Llama-3.1-8B-Instruct \cite{grattafiori2024llama3herdmodels} through word games, resulting in paprika-Meta-Llama-3.1-8B-Instruct. However, these approaches do not explicitly quantify the IG associated with each question, and therefore lack a fine-grained metric to assess LLM performance throughout the interaction. Moreover, the LLMs trained in these works do not leverage Chain-of-Thought (CoT) reasoning, leaving a gap in understanding how such strategies affect information-seeking performance in this setting.

\citet{mazzaccara2024learningaskinformativequestions} explore strategies to improve the informativeness of questions generated by the Llama-2-7b-chat model in the \emph{20-Questions}~\cite{MosherHornsby1966} game, employing Expected Information Gain (EIG) to guide question formulation. More generally, language games in which the underlying data are structured as explicit hypothesis spaces—represented as graphs whose leaf nodes correspond to possible target concepts—provide a natural setting for evaluating how effectively LLMs engage in information seeking and how well they can comprehend and operate within a hypothesis space. In such settings, the IG obtained from each question can be quantified by measuring how many candidate nodes the agent successfully eliminates from the hypothesis space. \citet{bertolazzi-etal-2023-chatgpts} adopt this general strategy, also using \emph{20-Questions} game, leveraging datasets structured as knowledge graphs and computing EIG. Despite their contributions, both works share important limitations: they restrict their analyses to a single model each (GPT-3.5-turbo and Llama-2-7b-chat, respectively), do not compare a broader range of open-source LLMs and do not examine whether CoT would impact information-seeking performance in this setting, even though prior work has shown its benefits on symbolic reasoning tasks~\cite{wei2023chainofthoughtpromptingelicitsreasoning}.

\section{Method}

This section describes the proposed system's architecture and the interactive agents comprising it.

\subsection{Environment}
The system operates within a hierarchical knowledge graph environment, structured across distinct abstraction levels. This graph conceptually forms a tree, where the terminal leaf nodes delineate all feasible targets within the operational domain. Each successive upper-layer node serves to aggregate and abstract the nodes from the immediate lower layer, based on a similarity criterion of the defined datasets considering specific, pertinent characteristics.

\subsection{Agents}

Our system employs three LLM-based agents that interact collaboratively within the hierarchical knowledge graph environment. Each agent fulfills a distinct role in the information-gathering process:

\begin{itemize}
    \item \textbf{Seeker:} The primary evaluation subject of our framework, the Seeker agent aims to identify the target node through a series of yes/no clarification questions. This agent's effectiveness in formulating informative queries that efficiently narrow down the search space is the central metric we aim to measure. The Seeker operates by analyzing the current state of the environment and formulating questions that maximize IG. 
    The Seeker operates under two distinct modes of observability:
    \begin{itemize}
        \item \textbf{Fully Observable (FO):} In this mode, at each turn, the Seeker has access to the complete current state of the knowledge graph along with the full history of previous turns in the conversation. This setting explicitly allows the Seeker to reason about the entire hypothesis space and make informed decisions about the next question to ask.
        \item \textbf{Partially Observable (PO):} Here, the Seeker’s perception is limited to the sequence of prior conversational turns only, without direct access to the present structure of the knowledge graph. This constrains the agent’s reasoning to only the information revealed during the interaction, requiring it to implicitly infer the hypothesis space.
    \end{itemize}

    \item \textbf{Oracle:} This agent possesses complete knowledge of the target node and serves as an omniscient responder. Upon receiving a yes/no question from the Seeker, the Oracle provides a definitive answer based on its knowledge of the true target. Additionally, the Oracle evaluates whether the Seeker has successfully identified the target node, thereby determining whether the game should continue to a subsequent turn or terminate.
    
    \item \textbf{Pruner:} The Pruner agent is responsible for dynamically refining the search space by eliminating infeasible nodes from the knowledge graph. After each question-answer exchange, the Pruner analyzes the current state of the graph and the information revealed by the Q\&A interaction, then determines which nodes should be pruned based on logical constraints derived from the Oracle's responses. For example, if the Seeker asks \emph{``Is the target in Asia?''} and the Oracle answers \emph{``No''}, the Pruner removes from the active hypothesis space all city nodes whose region or subregion is labeled as Asia, while keeping all remaining cities active.
\end{itemize}

\subsection{Game Protocol}

\begin{figure}[t]
\centering
\includegraphics[width=0.3\textwidth]{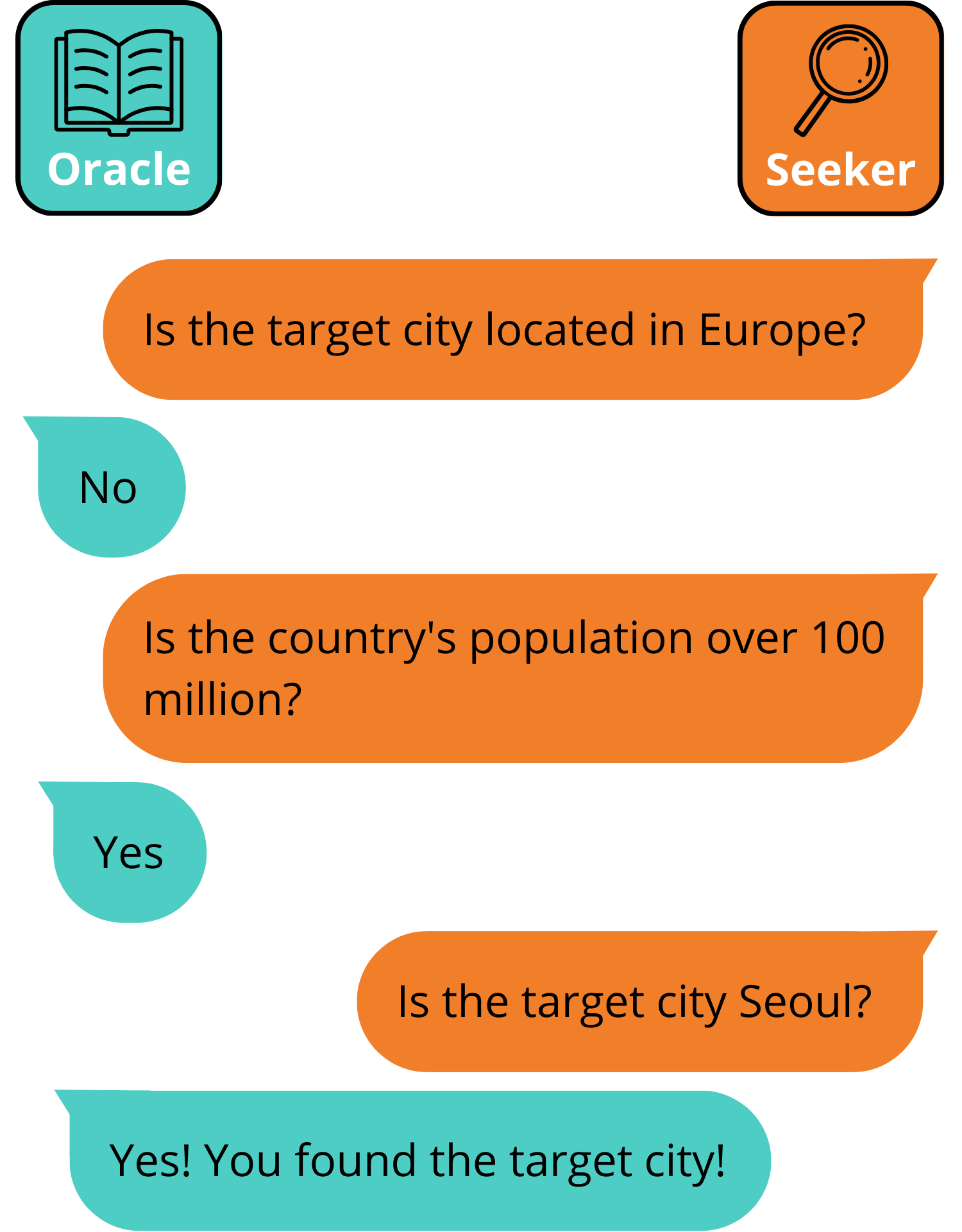}
\caption{\textbf{Example dialogue between the Seeker and Oracle agents.} The Seeker asks yes/no questions to narrow down the hypothesis space, while the Oracle provides answers based on its knowledge of the target. The game terminates when the Seeker identifies the target.}
\label{fig:conversation}
\end{figure}

The interaction between agents follows a structured turn-based protocol (exemplified in Figure~\ref{fig:conversation}), where each turn consists of three sequential phases:

\begin{enumerate}
    \item \textbf{Seeker's Phase:} The Seeker receives information about the environment state through the complete history of previous conversational turns (i.e., all prior questions and answers). In the FO setting, the Seeker additionally receives the current state of the knowledge graph. Based on this information, the Seeker formulates a yes/no question directed at the Oracle, designed to maximize IG and reduce the set of candidate target nodes. See Appendix~\ref{apx:seeker_prompt} for prompt.
    
    \item \textbf{Oracle's Phase:} The Oracle receives the Seeker's question and responds with a definitive yes or no answer based on its knowledge of the true target node. Following its response, the Oracle evaluates whether the Seeker has successfully identified the target. If the target has been correctly discovered, the game ends right after the Pruner's Phase.
    
    \item \textbf{Pruner's Phase:} The Pruner analyzes the Seeker's current question and the Oracle's response, together with the previous state of the knowledge graph. Based on this information, the Pruner applies the logical constraints to update the graph by pruning nodes that are no longer feasible candidates for the target.
    If any node was pruned, the IG is calculated.
\end{enumerate}

This iterative process continues until either the Seeker successfully identifies the target node or a predefined maximum number of turns is reached, at which point the game terminates.

\subsection{Information Gain as a Metric for Seeker Efficiency}

A central objective of our framework is to quantitatively assess the effectiveness of the Seeker agent's queries in reducing uncertainty during the information-gathering process. To this end, we employ \emph{IG} as the main metric, grounded in Shannon entropy.

\paragraph{Quantifying uncertainty.} At any given turn, the uncertainty regarding the target node is characterized by the set of candidate nodes that remain feasible hypotheses. Assuming a uniform prior—where each of the $N$ active nodes has an equal probability $p = 1/N$ of being the target—the Shannon entropy provides a natural quantification of this uncertainty:
\begin{equation}
    H = \log_2 N
\end{equation}
where $H$ represents the entropy in bits, and $N$ denotes the current number of active candidate nodes.

\paragraph{Definition of Information Gain.} The \emph{IG} at each turn is defined as the reduction in entropy resulting from the Seeker's question, the Oracle's answer, and the subsequent pruning step:
\begin{equation}
    IG = H_{\mathrm{before}} - H_{\mathrm{after}}
\end{equation}
where \( IG \) is measured in bits, \( H_{\mathrm{before}} \) is the entropy prior to the interaction, and \( H_{\mathrm{after}} \) is the entropy following application of all constraints.

High IG per turn indicates efficient and informative question-asking, while lower values signify uninformative or redundant interactions. This metric provides a direct, quantitative framework for evaluating and comparing the performance of LLM-based Seeker agents in structured information-seeking tasks.

\paragraph{Note on the generalizability of the framework.} By design, the framework presented could be applied to diverse language-based tasks. For example, it can be instantiated for a range of language games, including ``20 Questions'' and others proposed in the LMRL Gym benchmark~\cite{abdulhai2023lmrlgymbenchmarksmultiturn}, and could accommodate any domain in which data can be structured as known hypothesis spaces with enumerated options.



\section{Experimental Setup}

We present our experimental setup, dataset construction, model configurations, and evaluation protocol for assessing information-gathering capabilities in LLMs.

\subsection{Experimental Domain}

We instantiate our framework using geographical data, inspired by the \emph{Guess My City} game. This domain naturally provides a hierarchical structure that aligns with our knowledge graph architecture and enables effective evaluation of clarification question formulation.

\subsection{Dataset}

We construct our evaluation dataset from the Countries-States-Cities database (see Appendix~\ref{apx:dataset_structure}). We restrict the candidate set to the 40 most populous cities worldwide to ensure that even relatively small models (used as an oracle; 8B parameters) have sufficient world knowledge to answer the Seeker's questions about target entities. This choice reduces confounds from oracle knowledge gaps on long-tail or obscure cities, while keeping the hypothesis space non-trivial.

Our dataset includes the following attributes for each city: 
\begin{itemize}
    \item \textbf{City:} \texttt{city\_id}, \texttt{city\_name}
    \item \textbf{State:} \texttt{state\_id}, \texttt{state\_name}
    \item \textbf{Country:} \texttt{country\_id}, \texttt{country\_name}
    \item \textbf{Region:} \texttt{region\_id}, \texttt{region\_name}
    \item \textbf{Subregion:} \texttt{subregion\_id}, \texttt{subregion\_name}
\end{itemize}

We organize the hierarchical knowledge graph into a five-level taxonomy: \texttt{region} $\rightarrow$ \texttt{subregion} $\rightarrow$ \texttt{country} $\rightarrow$ \texttt{state} $\rightarrow$ \texttt{city}, where each level represents increasing geographical specificity.

\subsection{Experimental Protocol}

To ensure that our experiment is not biased by potential differences in difficulty across target cities, we use each of the 40 cities as a target node. Given the stochastic nature of LLM responses, we perform three independent runs for each city. This design yields 120 game instances (40 cities $\times$ 3 runs) for each model evaluated.

We set the maximum number of turns per game to 30, providing sufficient opportunity for the Seeker to identify the target while maintaining computational tractability. Games terminate when the Seeker successfully identifies the target node or when the maximum turn limit is reached. An example dialogue illustrating the interaction between the Seeker and Oracle agents is shown in Figure~\ref{fig:conversation}.

\subsection{Model Configuration}

\subsubsection{Oracle and Pruner Agents}

We implement both the Oracle and Pruner agents using Qwen3-8B for all experiments. This model is chosen to ensure: (1) sufficient reasoning capabilities for the Pruner to interpret the graph structure and logically prune nodes; (2) adequate knowledge for the Oracle to reliably answer questions about target cities; (3) reproducibility through the use of an open-source model; and (4) a strong balance between performance and computational cost. Notably, since the interactions involve objective factual queries and deterministic logical constraints rather than subjective evaluations or preference-based rankings, this homogeneous setup is not expected to suffer from the self-preference biases often observed in LLM-as-a-judge frameworks~\citep{chen2025llms}.

\subsubsection{Seeker Agent}

We evaluate multiple LLM variants in the Seeker role to assess how different architectures, sizes, and capabilities influence information-gathering performance:

\begin{itemize}
    \item \textbf{Llama-3.1-8B-Instruct}: A standard instruction-tuned model serving as our baseline.
    \item \textbf{Qwen3-30B-A3B-Thinking-2507}: A larger model with explicit reasoning capabilities.
    \item \textbf{Qwen3-8B}: A model matching the architecture and size of the Oracle and Pruner, enabling controlled comparisons.
    \item \textbf{paprika-Meta-Llama-3.1-8B-Instruct}: A fine-tuned variant of Llama-3.1-8B-Instruct using language games dataset \cite{tajwar2025traininggenerallycuriousagent}, allowing us to examine the impact of domain-specific adaptation.
\end{itemize}

\subsubsection{Observability Conditions}

For each Seeker model, we evaluate performance under two observability settings:

\begin{itemize}
    \item \textbf{Fully Observable (FO)}: The Seeker has complete access to the current state of the knowledge graph and the full interaction history at each turn.
    \item \textbf{Partially Observable (PO)}: The Seeker's perception is limited to the sequence of previous conversational turns, without direct access to the graph structure.
\end{itemize}

For models that support explicit reasoning capabilities (specifically, the Qwen variants), we conduct experiments with reasoning both enabled and disabled to investigate the impact of CoT reasoning on information-gathering effectiveness.

\begin{table*}[t]
\centering
\setlength{\tabcolsep}{1.0mm}
\small
\begin{tabular}{lcccccc}
\toprule
\textbf{Model} & \textbf{Obs.} & \textbf{CoT} & \textbf{Win Rate} $\uparrow$ & \textbf{Avg Turns} $\downarrow$ & \textbf{IG/Turn} $\uparrow$ & \textbf{Total IG} $\uparrow$ \\
\midrule
\multicolumn{7}{l}{\textbf{Llama-3.1-8B-Instruct}} \\
    & FO & No & $1.00 \pm 0.00$ & $9.33 \pm 0.43$ & $0.64 \pm 0.03$ & $5.32 \pm 0.00$ \\
    & PO & No & $0.81 \pm 0.04$ & $18.65 \pm 0.96$ & $0.35 \pm 0.02$ & $5.21 \pm 0.05$ \\
\midrule
\multicolumn{7}{l}{\textbf{Qwen3-30B-A3B-Instruct-2507}} \\
    & FO & No & $1.00 \pm 0.00$ & $7.95 \pm 0.23$ & $0.70 \pm 0.02$ & $5.32 \pm 0.00$ \\
    & PO & No & $0.10 \pm 0.04$ & $29.13 \pm 0.49$ & $0.14 \pm 0.01$ & $3.81 \pm 0.22$ \\
\midrule
\multicolumn{7}{l}{\textbf{Qwen3-30B-A3B-Thinking-2507}} \\
    & FO & Yes & $1.00 \pm 0.00$ & $7.12 \pm 0.15$ & $0.78 \pm 0.02$ & $5.32 \pm 0.00$ \\
    & PO & Yes & $0.93 \pm 0.03$ & $14.88 \pm 0.78$ & $0.42 \pm 0.02$ & $5.11 \pm 0.10$ \\
\midrule
\multicolumn{7}{l}{\textbf{Qwen3-8B}} \\
    & FO & Yes & $1.00 \pm 0.00$ & $8.86 \pm 0.33$ & $0.65 \pm 0.03$ & $5.32 \pm 0.00$ \\
    & FO & No & $0.77 \pm 0.04$ & $16.02 \pm 0.83$ & $0.42 \pm 0.03$ & $4.80 \pm 0.12$ \\
    & PO & Yes & $0.94 \pm 0.03$ & $15.82 \pm 0.74$ & $0.41 \pm 0.03$ & $5.32 \pm 0.00$ \\
    & PO & No & $0.08 \pm 0.03$ & $29.60 \pm 0.20$ & $0.17 \pm 0.00$ & $4.95 \pm 0.10$ \\
\midrule
\multicolumn{7}{l}{\textbf{paprika-Meta-Llama-3.1-8B-Instruct}} \\
    & FO & No & $1.00 \pm 0.00$ & $9.57 \pm 0.52$ & $0.63 \pm 0.03$ & $5.32 \pm 0.00$ \\
    & PO & No & $0.78 \pm 0.04$ & $17.18 \pm 1.00$ & $0.40 \pm 0.03$ & $5.14 \pm 0.06$ \\
\bottomrule
\end{tabular}
\caption{\textbf{Performance comparison across Seeker models.} Models with better reasoning 
capabilities consistently achieve higher 
information gain and success rates, while requiring 
fewer turns. Results are reported as mean $\pm$ standard error over 40 cities (3 runs each). Obs.: observability (FO = Fully Observable, PO = Partially Observable). CoT: Chain-of-Thought reasoning enabled. Win Rate: fraction of successful target identifications. Avg Turns: turns per game. IG/Turn: information gain per turn. Total IG: cumulative information gain. Arrows indicate preferred direction.}
\label{tab:results}
\end{table*}

\section{Results}

\begin{figure*}[t]
    \centering
    \includegraphics[width=1\linewidth]{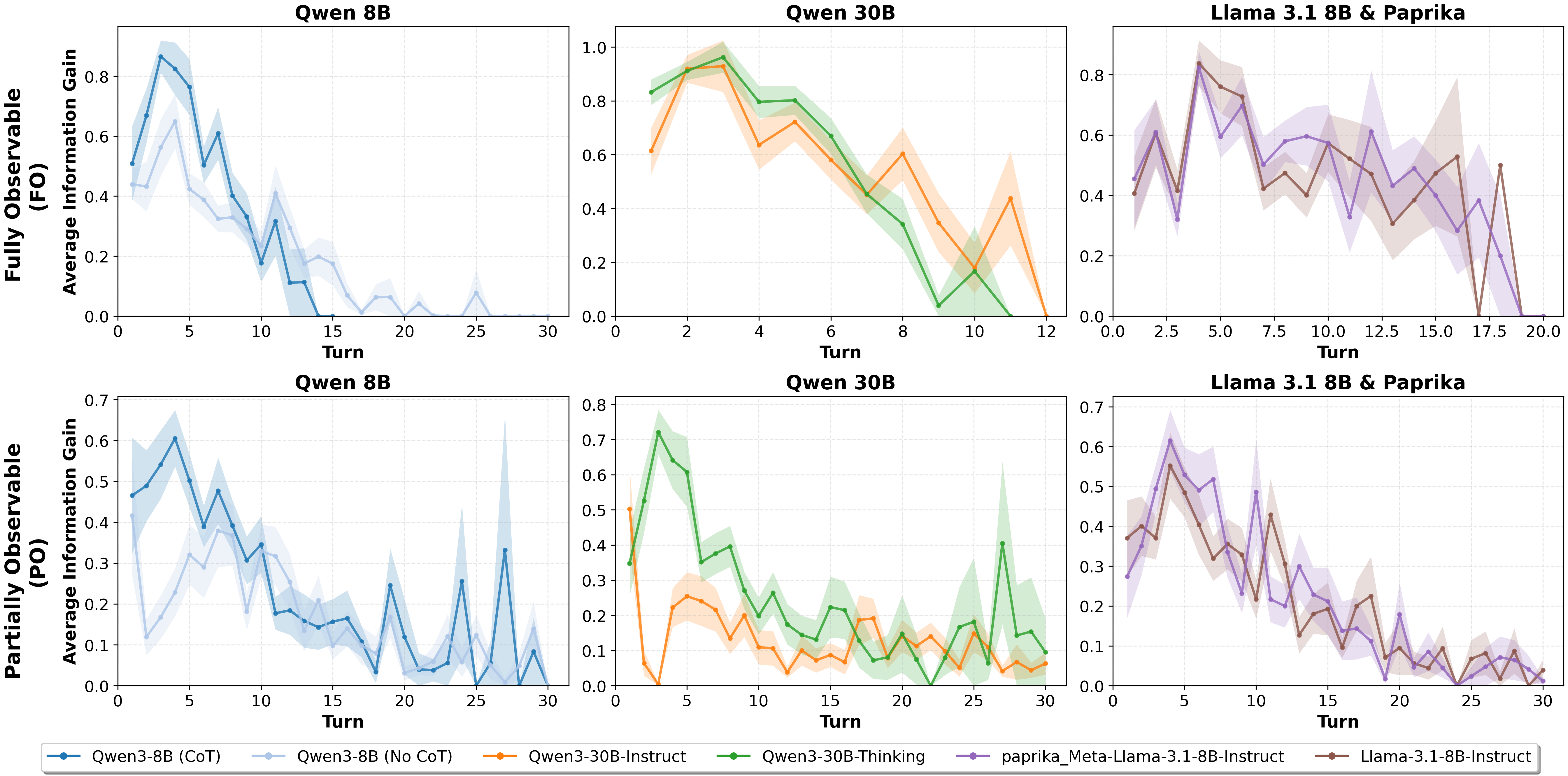}
    \caption{\textbf{Aggregated Information Gain Over Time.} Performance comparison of LLMs under fully observable (FO) and partially observable (PO) conditions. Each subplot reports average information gain per turn. Models with stronger reasoning yield higher information gain, especially early; performance drops as the hypothesis space narrows.}
    \label{fig:aggregated_ig}
\end{figure*} 

We present experimental results evaluating the information-gathering capabilities of different Seeker models across observability conditions and reasoning modes. Table~\ref{tab:results} summarizes our findings, while Figure~\ref{fig:aggregated_ig} visualizes the information gain trends.

As shown in Table~\ref{tab:results}, in the PO setting, models equipped with explicit reasoning capabilities demonstrate a superior ability to navigate the hypothesis space, formulating queries that yield higher average IG. This is particularly clear for the Qwen3-8B and Qwen3-30B variants, where activating CoT reasoning results in marked improvements in both turn efficiency and IG.

For instance, the Qwen3-8B model, when evaluated without CoT, exhibits a decline in its ability to implicitly navigate the hypothesis space and reach the correct solution. However, enabling CoT enhances its performance, allowing it to outperform both Llama-3.1-8B-Instruct and paprika-Meta-Llama-3.1-8B-Instruct in win rate, while achieving comparable or slightly higher IG per turn.

It is notable that, in the PO setting, the paprika-Meta-Llama-3.1-8B-Instruct model shows a slightly higher IG per turn. This suggests that the additional post-training on language games may have contributed to more inquisitive question-asking behavior, enabling it to generate questions that reduce entropy more efficiently during information-seeking tasks, though the effects are small.

\begin{table*}[t]
\centering
\setlength{\tabcolsep}{1.0mm}
\small
\begin{tabular}{lccccc}
\toprule
\textbf{Model} & \textbf{Obs.} & \textbf{Avg Optimal Rate} $\uparrow$ & \textbf{Avg Chosen IG} $\uparrow$ & \textbf{Avg Optimal IG} $\uparrow$ & \textbf{Avg Questions/Turn} $\uparrow$ \\
\midrule
\multicolumn{6}{l}{\textbf{Qwen3-30B-A3B-Thinking-2507}} \\
 & FO & $0.26 \pm 0.16$ & $0.77 \pm 0.21$ & $1.40 \pm 0.41$ & $5.93 \pm 1.24$ \\
 & PO & $0.25 \pm 0.13$ & $0.43 \pm 0.24$ & $0.88 \pm 0.34$ & $7.63 \pm 2.15$ \\
\midrule
\multicolumn{6}{l}{\textbf{Qwen3-8B}} \\
 & FO & $0.36 \pm 0.17$ & $0.66 \pm 0.22$ & $1.11 \pm 0.28$ & $5.05 \pm 1.22$ \\
 & PO & $0.19 \pm 0.11$ & $0.41 \pm 0.29$ & $0.94 \pm 0.49$ & $9.74 \pm 2.31$ \\
\bottomrule
\end{tabular}
\caption{\textbf{Decision quality during Seeker Chain-of-Thought.} Larger models generate candidates with higher potential IG, while smaller models compensate through broader exploration. Values are mean $\pm$ std. FO = Fully Observable; PO = Partially Observable. Avg Optimal Rate: proportion of turns selecting the maximum-IG candidate. Avg Chosen/Optimal IG: mean IG of executed/best candidates (bits). Avg Questions/Turn: candidates considered per turn.}
\label{tab:optimal_choice}
\end{table*}

\subsection{Analysis of Decision Quality}

To better understand the mechanisms underlying CoT that contribute to improved efficiency, we conducted an additional experiment examining the decision-making process of models equipped with explicit reasoning capabilities. For the CoT-enabled variants (Qwen3-8B and Qwen3-30B-A3B-Thinking-2507), we extracted all candidate questions that the Seeker agent considered during its reasoning process before selecting the final question. Using Gemini 2.5 Flash~\cite{comanici2025gemini25pushingfrontier}, we parsed the reasoning traces to identify these candidate questions explicitly mentioned in the model's reasoning. For each candidate question, we computed the IG it would have generated if it had been selected and executed. This analysis enables us to compare the IG of the questions actually chosen by the agent against the IG of alternative questions that were considered but not selected, providing insights into the decision quality and information-seeking strategies of CoT-enabled models. The results of this analysis, including optimal choice rates and IG comparisons, are presented in Table~\ref{tab:optimal_choice}.

In the PO setting (Table~\ref{tab:optimal_choice}), Qwen3-8B explores a larger set of candidate questions per turn than Qwen3-30B-A3B-Thinking-2507 (Avg Questions/Turn 9.74 vs.\ 7.63), but selects the optimal candidate less frequently (Avg Optimal Rate 0.19 vs.\ 0.25). This trade-off between broader exploration and a lower optimal-selection rate helps explain why the two models achieve similar Total IG and win rates in the PO–CoT condition (Table~\ref{tab:results}), despite Qwen3-30B-A3B-Thinking-2507 being considerably larger: the 8B model partially compensates for its smaller capacity by exploring more aggressively.

In the FO setting, both models are naturally discouraged from extensive exploration, likely because the hypothesis space is already made explicit by the graph provided in the prompt. Here, Qwen3-8B exhibits a higher Avg Optimal Rate than Qwen3-30B-A3B-Thinking-2507 (0.36 vs.\ 0.26), indicating that it more often selects the best candidate among those it considers. However, the 30B model attains higher Avg Chosen IG and Avg Optimal IG (0.77 vs.\ 0.67 and 1.39 vs.\ 1.11, respectively), suggesting that, even when it does not always pick the optimal candidate, it tends to generate candidate questions with higher potential IG overall.

\section{Conclusion}

We present a quantitative methodology for measuring the Information Gain of yes/no queries generated by LLMs. We instantiate this framework in the \emph{Guess My City} game and systematically evaluate open LLMs from different model families and sizes, with and without Chain-of-Thought reasoning, under both fully and partially observable conditions. We also conduct a fine-grained analysis of reasoning traces to characterize how models of different scales explore the hypothesis space and select informative questions. The methodology supports evaluation beyond standard metrics such as win rate or average number of turns; specifically, it enables (i) assessment of cumulative performance across multi-turn interactions and (ii) turn-level quantification of the informativeness of individual queries. In our experiments, larger models, especially when equipped with CoT, achieve higher IG and solve tasks in fewer steps, indicating a more effective exploitation of the underlying hypothesis structure. Finally, the reasoning-trace analysis reveals that larger models tend to generate candidate questions with higher potential IG, while smaller models partially compensate through broader exploration with more candidates per turn.

\paragraph{Limitations.}

First, the framework requires prior knowledge of the complete hypothesis space, which must be explicitly structured as a hierarchical knowledge graph. 
Second, our empirical evaluation focuses on the \emph{Guess My City}, a geographical game, which may not be representative of all possible domains. 
Third, the effectiveness of the framework depends on the knowledge and reasoning capabilities of the LLMs used as Oracle and Pruner agents, which must possess sufficiently accurate information about the target and candidate nodes to perform their tasks reliably.

\paragraph{Future work.}
Future work should investigate the framework's effectiveness across diverse domains and language games beyond the geographical setting considered. Moreover, enriching the hypothesis space with additional details could help reduce the framework's reliance on pre-existing model knowledge. Finally, the turn-level and cumulative IG metrics could be explored as reward signals in reinforcement learning, encouraging LLM-based agents to select questions that maximize expected Information Gain during multi-turn interactions.

\bibliography{aaai2026}

\onecolumn
\appendix

\section{Agent System Prompts}\label{apx:seeker_prompt}

This appendix provides the complete system prompts used for each of the three LLM-based agents in the framework: the Seeker, Oracle, and Pruner agents.

\subsection{Seeker Agent Prompt}

The Seeker agent is responsible for formulating yes/no questions to identify the target node. The system prompt instructs the agent to ask strategic questions that maximize information gain:

\begin{lstlisting}[language={},basicstyle={\footnotesize\ttfamily},numbers=none,frame=single,breaklines=true,breakatwhitespace=true,lineskip=2pt,aboveskip=6pt,belowskip=6pt]
# SeekerAgent System Prompt

## The Game

You are playing a geographic guessing game where your goal is to discover a secret target city through strategic questions.

### Players:
- You (Seeker): Ask yes/no questions to find the target
- Oracle: Knows the target and answers your questions truthfully
- Computer: Tracks the game state and remaining possibilities

### Your Role:
You are the Seeker - an intelligent detective trying to identify a specific target city in a geographic knowledge graph. Your goal is to ask strategic yes/no questions that maximize information gain and eliminate possibilities efficiently.

## Game Rules

1. Ask ONLY yes/no questions
2. Be specific and clear
3. Aim to eliminate roughly half the remaining possibilities
4. Avoid questions that reveal the specific target
5. Return only the question text, no explanations
\end{lstlisting}

\subsection{Oracle Agent Prompt}

The Oracle agent possesses complete knowledge of the target node and provides truthful yes/no answers to the Seeker's questions:

\begin{lstlisting}[language={},basicstyle={\footnotesize\ttfamily},numbers=none,frame=single,breaklines=true,breakatwhitespace=true,lineskip=2pt,aboveskip=6pt,belowskip=6pt]
# OracleAgent System Prompt

## The Game

You are playing a geographic guessing game where a Seeker tries to discover a secret target city through strategic questions.

### Your Role:
You are the Oracle - the all-knowing guide who possesses secret knowledge about the target location. Your role is to help the Seeker discover the target through truthful answers while maintaining the challenge and never revealing the target directly.

## Game Rules

1. Answer with simple "Yes" or "No"
2. Be truthful - never lie about the target's properties
3. NEVER reveal the target's name or ID directly
4. Keep answers brief and focused
5. If the question is unclear, ask for clarification
6. If you cannot answer with yes/no, provide minimal helpful information
7. The target is always a city
8. CRITICAL: Detect when the Seeker has found the target city, saying it's name, and end the game

## Response Format

You MUST respond with a JSON object containing these keys IN THIS ORDER:
1. rationale: Brief internal reasoning (1 sentence, not shown to Seeker)
2. answer: Your response to the Seeker (string)
3. game_over: Whether the Seeker has found the target (boolean)
\end{lstlisting}

\subsection{Pruner Agent Prompt}

The Pruner agent dynamically refines the search space by eliminating infeasible nodes based on logical constraints derived from the question-answer interactions:

\begin{lstlisting}[language={},basicstyle={\footnotesize\ttfamily},numbers=none,frame=single,breaklines=true,breakatwhitespace=true,lineskip=2pt,aboveskip=6pt,belowskip=6pt]
You are the PrunerAgent for a knowledge-graph benchmark.

Goal:
- Given the current graph state (in text), the turn index, and the last Q&A, decide which CITY node IDs to prune. Only prune when logically implied by the question and answer. Prefer minimal, conservative pruning.

Rules:
- Never reveal or assume the hidden target.
- Consider only ACTIVE nodes in the provided graph text.
- CRITICAL: ONLY CITY NODES CAN BE TARGETS
- CRITICAL PRUNING LOGIC:
  * If answer is "No" to "Is target in X?", prune ONLY CITY nodes that ARE in X
  * If answer is "Yes" to "Is target in X?", prune ONLY CITY nodes that are NOT in X
  * Example: Q="Is target in North America?" A="No" -> Prune CITY nodes IN North America, KEEP all others
  * Example: Q="Is target in Asia?" A="Yes" -> Prune CITY nodes NOT in Asia, KEEP Asian CITY nodes
  * NEVER prune countries, states, regions, or subregions - only cities
- If ambiguous, do not prune.

Output:
- Return ONLY a JSON object with exactly two keys IN THIS ORDER:
  {"rationale": "short explanation", "pruned_ids": ["city:id1", "city:id2", ...]}
- Do not include any extra commentary or formatting.
- pruned_ids must contain ONLY city IDs (starting with "city:")
\end{lstlisting}

\clearpage

\section{Experimental Configuration Details}

This section provides detailed configuration parameters and hyperparameters used in the experiments to ensure reproducibility.

\subsection{Model Hyperparameters}

Each model in the experiments was used with its default hyperparameter configuration as specified by the respective provider.

\subsection{Dataset Structure}\label{apx:dataset_structure}

The evaluation dataset is constructed by combining two data sources:

\begin{enumerate}
    \item \textbf{Countries-States-Cities Database}\footnote{Available at: \url{https://github.com/dr5hn/countries-states-cities-database}}: Provides the hierarchical geographical structure (regions, subregions, countries, states, and cities) with identifiers and names for each level.
    \item \textbf{World Population Review}\footnote{Available at: \url{https://worldpopulationreview.com/cities}}: Provides 2025 population estimates for cities worldwide, used to identify and rank the most populous cities.
\end{enumerate}

\paragraph{Dataset construction process.} The evaluation dataset was constructed through the following steps:

\begin{enumerate}
    \item \textbf{Data Loading:} The hierarchical geographical data was loaded from the Countries-States-Cities database JSON files (\texttt{regions.json}, \texttt{subregions.json}, \texttt{countries.json}, \texttt{states.json}, and \texttt{cities.json}).
    \item \textbf{Data Integration:} The hierarchical data structures (cities $\rightarrow$ states $\rightarrow$ countries $\rightarrow$ regions $\rightarrow$ subregions) were merged to create a flat representation with complete geographical context for each city.
    \item \textbf{Population Data Integration:} The integrated geographical data was merged with 2025 population estimates from World Population Review, matching cities by name and country.
    \item \textbf{Population-Based Filtering:} The merged dataset was filtered to include only the top 40 most populous cities worldwide based on 2025 population estimates, ensuring that all evaluated LLMs have sufficient knowledge about the target entities.
    \item \textbf{CSV Export:} The final filtered dataset was exported to CSV format for use in the experiments.
\end{enumerate}

\paragraph{Final dataset format.} The resulting dataset is stored as \texttt{data/top\_40\_pop\_cities.csv} and contains the following attributes for each of the 40 cities:

\begin{itemize}
    \item \textbf{File format:} CSV (Comma-Separated Values)
    \item \textbf{Location:} \texttt{data/top\_40\_pop\_cities.csv}
    \item \textbf{Filtering criterion:} Top 40 most populous cities worldwide (based on 2025 population estimates)
    \item \textbf{Attributes per city:} \texttt{city\_id}, \texttt{city\_name}, \texttt{state\_id}, \texttt{state\_name}, \texttt{country\_id}, \texttt{country\_name}, \texttt{region\_id}, \texttt{region\_name}, \texttt{subregion\_id}, \texttt{subregion\_name}, \texttt{population\_2025}
\end{itemize}

This hierarchical structure enables the construction of a five-level knowledge graph taxonomy: \texttt{region} $\rightarrow$ \texttt{subregion} $\rightarrow$ \texttt{country} $\rightarrow$ \texttt{state} $\rightarrow$ \texttt{city}, where each level represents increasing geographical specificity.

\subsection{Experimental Protocol Configuration}

The experimental protocol follows these specifications:

\begin{itemize}
    \item \textbf{Total targets:} 40 cities (all cities in the filtered dataset)
    \item \textbf{Runs per target:} 3 independent runs per city
    \item \textbf{Total game instances:} 120 (40 cities $\times$ 3 runs)
    \item \textbf{Maximum turns per game:} 30
    \item \textbf{Termination conditions:} Game ends when (1) Seeker successfully identifies the target, or (2) maximum turn limit is reached
\end{itemize}

\subsection{Optimal Choice Analysis}

Table~\ref{tab:optimal_choice} presents an analysis of the decision quality of CoT-enabled Seeker agents by examining the candidate questions they consider during reasoning. For each turn, we compare the IG of questions actually selected by the agent against the IG of alternative candidate questions that were considered but not chosen, quantifying how frequently the agent selects the optimal question from its reasoning process.

\clearpage


\end{document}